\documentclass[11pt,a4paper]{article}
\usepackage[hyperref]{acl2021}
\usepackage{times}
\usepackage{latexsym}

\usepackage{microtype}
\usepackage{times}
\usepackage{latexsym}
\usepackage{graphicx}  
\usepackage{subfigure}
\usepackage{url}
\usepackage{amsfonts} 
\usepackage{amsmath}
\usepackage{booktabs}
\usepackage{multirow}
\usepackage{arydshln}
\usepackage{booktabs}
\usepackage{pgfplots}
\usepackage{subfigure}

\definecolor{orange1}{RGB}{255, 145, 78}
\definecolor{blue1}{RGB}{88, 188, 255}
\usepackage{pifont}
\aclfinalcopy 

\title{Structural Pre-training for Dialogue Comprehension}

\author{
   Zhuosheng Zhang\textsuperscript{1,2,3},
	Hai Zhao\textsuperscript{1,2,3,\thanks{\; Corresponding author.  This paper was partially supported by National Key Research and Development Program of China (No. 2017YFB0304100), Key Projects of National Natural Science Foundation of China (U1836222 and 61733011), Huawei-SJTU long term AI project, Cutting-edge Machine reading comprehension and language model.}}\\

    \textsuperscript{\rm 1}Department of Computer Science and Engineering, Shanghai Jiao Tong University\\
	\textsuperscript{\rm 2}Key Laboratory of Shanghai Education Commission for Intelligent Interaction\\
	and Cognitive Engineering, Shanghai Jiao Tong University, Shanghai, China\\
	\textsuperscript{\rm 3}MoE Key Lab of Artificial Intelligence, AI Institute, Shanghai Jiao Tong University\\
	{\tt
	zhangzs@sjtu.edu.cn, zhaohai@cs.sjtu.edu.cn}
}

\date{}

\begin{document}
\maketitle
\begin{abstract}
Pre-trained language models (PrLMs) have demonstrated superior performance due to their strong ability to learn universal language representations from self-supervised pre-training. However, even with the help of the powerful PrLMs, it is still challenging to effectively capture task-related knowledge from dialogue texts which are enriched by correlations among speaker-aware utterances. In this work, we present SPIDER, Structural Pre-traIned DialoguE Reader, to capture dialogue exclusive features. To simulate the dialogue-like features, we propose two training objectives in addition to the original LM objectives: 1) utterance order restoration, which predicts the order of the permuted utterances in dialogue context; 2) sentence backbone regularization, which regularizes the model to improve the factual correctness of summarized subject-verb-object triplets. Experimental results on widely used dialogue benchmarks verify the effectiveness of the newly introduced self-supervised tasks.

\end{abstract}
\section{Introduction}
Recent advances in large-scale pre-training language models (PrLMs) have achieved remarkable successes in a variety of natural language processing (NLP) tasks \cite{Peters2018ELMO,radford2018improving,devlin-etal-2019-bert,yang2019xlnet,clark2020electra}. Providing fine-grained contextualized embedding, these pre-trained models are widely employed as encoders for various downstream NLP tasks. Although the PrLMs demonstrate superior performance due to their strong representation ability from self-supervised pre-training, it is still challenging to effectively adapt task-related knowledge during the detailed task-specific training which is usually in a way of fine-tuning \cite{gururangan-etal-2020-dont}. Generally, those PrLMs handle the whole input text as a linear sequence of successive tokens and implicitly capture the contextualized representations of those tokens through self-attention. Such fine-tuning paradigm of exploiting PrLMs would be suboptimal to model dialogue task which holds exclusive text features that plain text for PrLM training may hardly embody. Therefore, we explore a fundamental way to alleviate this difficulty by improving the training of PrLM. This work devotes itself to designing the natural way of adapting the language modeling to the dialogue scenario motivated by the natural characteristics of dialogue contexts.

\begin{figure}
\centering
\includegraphics[width=0.5\textwidth]{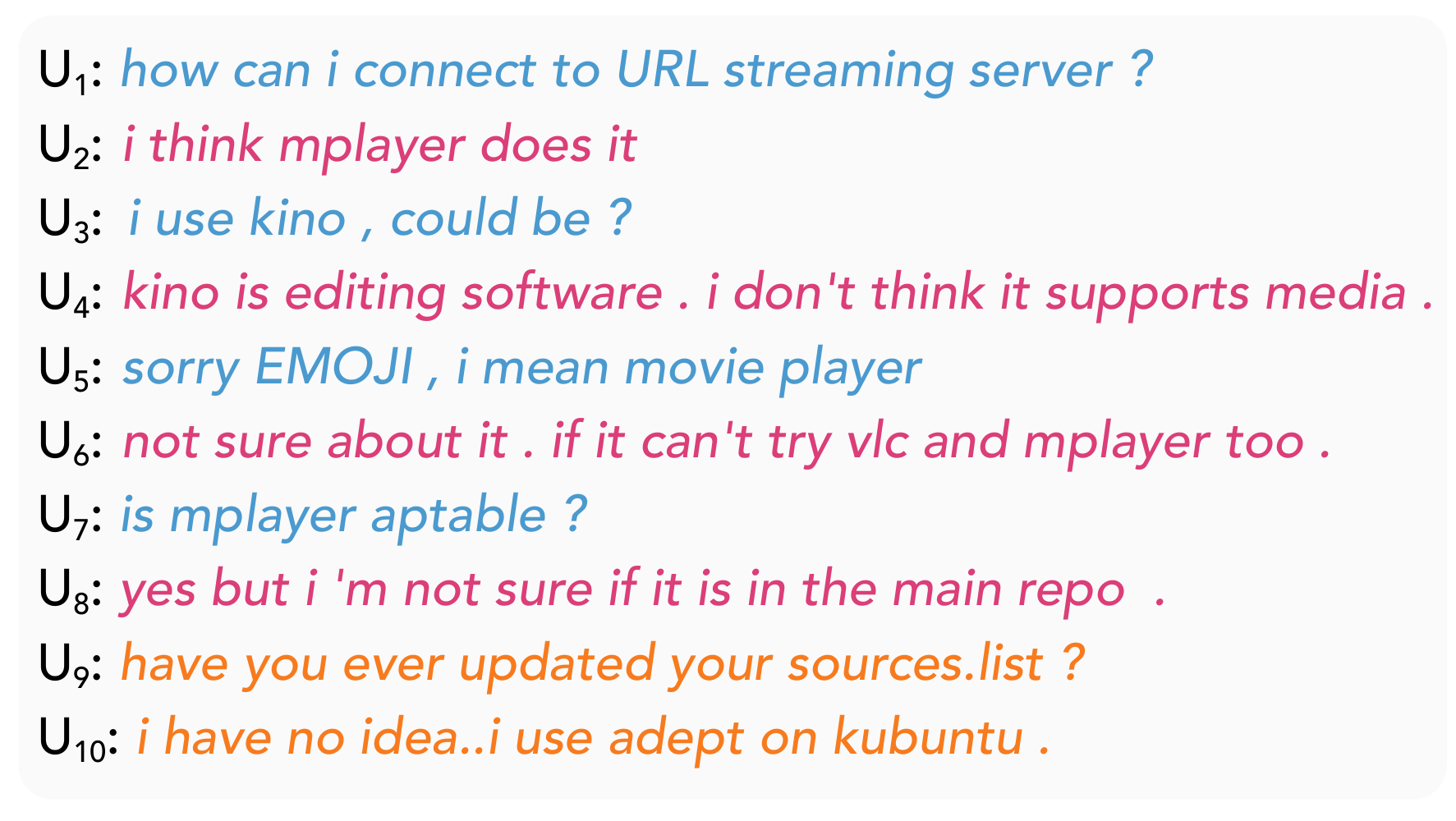}
\caption{A multi-turn dialogue example. Different colors indicate the utterances from different speakers.}
\label{dialogue_exp}
\end{figure}

As an active research topic in the NLP field, multi-turn dialogue modeling has attracted great interest. The typical task is response selection \cite{lowe2015ubuntu,wu2016sequential,zhang2018modeling} that aims to select the appropriate response according to a given dialogue context containing a number of utterances, which is the focus in this work. However, selecting a coherent and informative response for a given dialogue context remains a challenge. The multi-turn dialogue typically involves two or more speakers that engage in various conversation topics, intentions, thus the utterances are rich in interactions, e.g., with criss-cross discourse structures \cite{li-etal-2020-molweni}. A critical challenge is the learning of rich and robust context representations and interactive relationships of dialogue utterances, so that the resulting model is capable of adequately capturing the semantics of each utterance, and the relationships among all the utterances inside the dialogue.



Inspired by the effectiveness for learning universal language representations of PrLMs, there are increasing  studies that employ PrLMs for conversation modeling \cite{mehri2019pretraining,zhang2019dialogpt,rothe2020leveraging}. These studies typically model the response selection with only the context-response matching task and overlook many potential training signals contained in dialogue data. 
Although the PrLMs have learned contextualized semantic representation from token-level or
sentence-level pre-training tasks like MLM, NSP,
they all do not consider dialogue related features like speaker role, continuity and consistency.
One obvious issue of these approaches is that the relationships between utterances are harder to capture using word-level semantics. Besides, some latent features, such as user intent and conversation topic, are under-discovered in existing works \cite{xu2021topic}. 
Therefore, the response retrieved by existing dialogue systems supervised by the conventional way still faces critical challenges, including incoherence and inconsistency.




In this work, we present SPIDER (Structural Pre-traIned DialoguE Reader), a structural language modeling method to capture dialogue exclusive features. Motivated to efficiently and explicitly model the coherence among utterances and the key facts in each utterance, we propose two training objectives in analogy to the original BERT-like language model (LM) training: 1) utterance order restoration (UOR), which predicts the order of the permuted utterances in dialogue context; 2) sentence backbone regularization (SBR), which regularizes the model to improve the factual correctness of summarized subject-verb-object (SVO) triplets. Experimental results on widely used benchmarks show that SPDER boosts the model performance for various multi-turn dialogue comprehension tasks including response selection and dialogue reasoning.

\section{Background and Related Work}
\subsection{Pre-trained Language Models}
Recent works have explored various architecture choices and training objectives for large-scale LM pre-training. Most of the PrLMs are based on the encoder in Transformer, among which Bidirectional Encoder Representations from Transformers (BERT) \cite{devlin2018bert} is one of the most representative work. BERT uses multiple layers of stacked Transformer Encoder to obtain contextualized representations of the language at different levels. BERT has helped achieve great performance improvement in a broad range of NLP tasks. Several subsequent variants have been proposed to further enhance the capacity of PrLMs, such as XLNet \cite{yang2019xlnet}, RoBERTa \cite{liu2019roberta}, ALBERT \cite{lan2019albert}, ELECTRA \cite{clark2020electra}. For simplicity and convenient comparison with public studies, we select the most widely used BERT as the backbone in this work.

There are two ways of training PrLMs on dialogue scenarios, including open-domain pre-training and domain-adaptive post-training. Some studies perform training on open-domain conversational data
like Reddit for response selection or generation
tasks \cite{wolf2019transfertransfo,zhang2020dialogpt,henderson-etal-2020-convert,bao-etal-2020-plato}, but they are limited to the original pre-training
tasks and ignore the dialogue related features. For domain-adaptive post-training, prior works have indicated that the order information would be important in the text representation, and the well-known next-sentence-prediction \cite{devlin2018bert} and sentence-order-prediction \cite{lan2019albert} can be viewed as special cases of order prediction. Especially in the dialogue scenario, predicting the word order of utterance, as well as the utterance order in the context, has shown effectiveness in the dialogue generation task \cite{kumar2020deep,gu2020dialogbert}, where the order information is well recognized \cite{chen-etal-2019-neural}. However, there is little attention paid to dialogue comprehension tasks such as response selection \cite{lowe2015ubuntu,wu2016sequential,zhang2018modeling}. The potential difficulty is that utterance order restoration involves much more ordering possibilities for utterances that may have a quite flexible order inside dialogue text than NSP and SOP which only handle the predication of two-class ordering.

Our work is also profoundly related to auxiliary multi-task learning, whose common theme is to guide the language modeling Transformers with explicit knowledge and complementing objectives \cite{zhang-etal-2019-ernie,sun2019ernie,xu2020learning}. A most related work is \citet{xu2020learning}, which introduces four self-supervised tasks including next session prediction, utterance restoration, incoherence detection and consistency discrimination. Our work differs from \citet{xu2020learning} by three sides. 1) Motivation: our method is designed for a general-purpose in broad dialogue comprehension tasks whose goals may be either utterance-level discourse coherence or inner-utterance factual correctness, instead of only motivated for downstream context-response matching, whose goal is to measure if two sequences are related or not. 2) Technique: we propose both sides of intra- and inter- utterance objectives. In contrast, the four objectives proposed in \citet{xu2020learning} are natural variants of NSP in BERT, which are all utterance-level. 3)	Training: we empirically evaluate domain-adaptive training and multi-task learning, instead of only employing multi-task learning, which requires many efforts of optimizing coefficients in the loss functions, which would be time-consuming. 

In terms of factual backbone modeling, compared with the existing studies that enhance the PrLMs by annotating named entities or incorporating external knowledge graphs \citep{eric2017key,liu2018knowledge}, the SVO triplets extracted in our sentence backbone predication objective (SBP) method, appear more widely in the text itself. Such triplets ensure the correctness of SVO and enable our model to discover the salient facts from the lengthy texts, sensing the intuition of ``who did what". 

\subsection{Multi-turn Dialogue Comprehension}
\label{sec:relatedwork}
Multi-turn dialogue comprehension aims to teach machines to read dialogue contexts and solve tasks such as response selection \cite{lowe2015ubuntu,wu2016sequential,zhang2018modeling} and answering questions \cite{sun2019dream,cui2020mutual}, whose common application is building intelligent human-computer interactive systems \cite{Chen2017survey,Shum2018,AliMe,zhu2018lingke}.
Early studies mainly focus on the matching between the dialogue context and question \cite{huang2018flowqa,zhu2018sdnet}. Recently, inspired by the impressive performance of PrLMs, the mainstream is employing PrLMs to handle the whole input texts of context and question, as a linear sequence of successive tokens and implicitly capture the contextualized representations of those tokens through self-attention \cite{qu2019bert,liu2020hisbert}. Such a way of modeling would be suboptimal to capture the high-level relationships between utterances in the dialogue history. In this work, we are motivated to model the structural relationships between utterances from utterance order restoration and the factual correctness inside each utterance in the perspective of language modeling pre-training instead of heuristically stacking deeper model architectures.


\begin{figure*}
\centering
\includegraphics[width=1.0\textwidth]{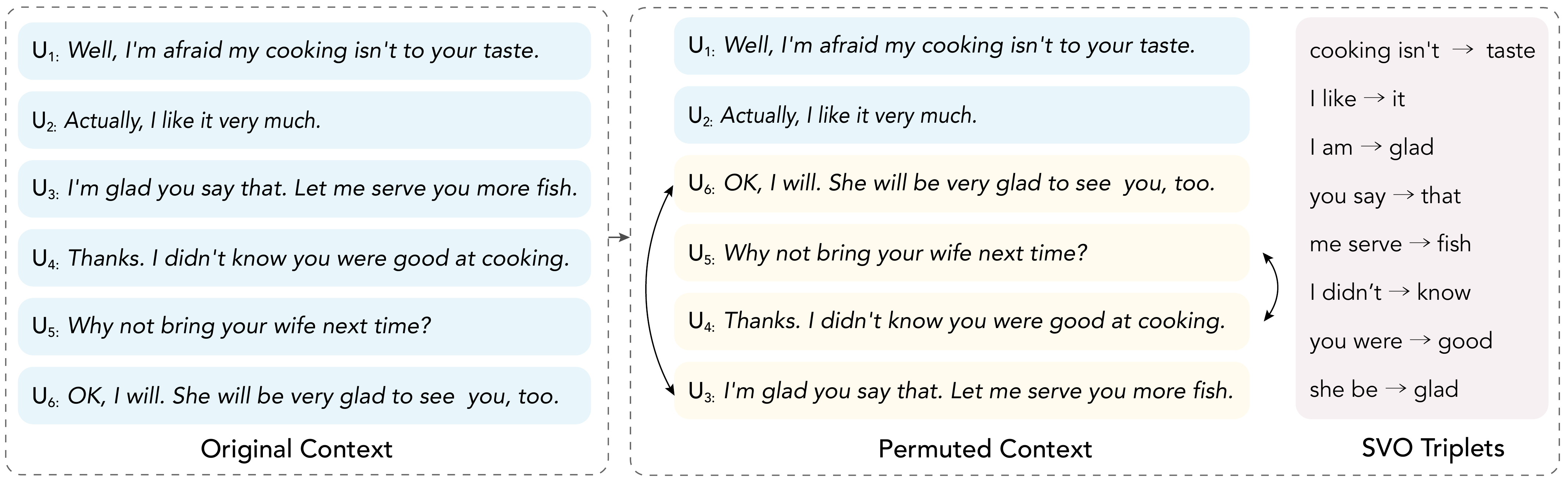}
\caption{Structural language modeling manipulations.}
\label{struct}
\end{figure*}

\section{Approach}
This section presents our proposed method SPIDER (Structural Pre-traIned DialoguE Reader). First, we will present the standard dialogue comprehension model as the backbone. Then, we will introduce our designed language modeling objectives for dialogue scenarios, including utterance order restoration (UOR) and sentence backbone regularization (SBR). In terms of model training, we employ two strategies, i.e., 1) domain adaptive post-training that first trains a language model based on newly proposed objectives and then fine-tunes the response selection task; 2) multi-task fine-tuning that trains the model for downstream tasks, along with LM objectives.
\subsection{Transformer Encoder}\label{sec:response_selection}
We first employ a pre-trained language model such as BERT \cite{devlin-etal-2019-bert} to obtain the initial word representations. The utterances and response are concatenated and then fed into the encoder. Given the context $C$ and response ${R}$, we concatenate all utterances in the context and the response candidate as a single consecutive token sequence with special tokens separating them: 
${X} = \{\texttt{[CLS]}  {R}  \texttt{[SEP]}  {U}_1 \texttt{[EOU]} \dots \texttt{[EOU]} {U}_n \texttt{[SEP]}\}$,
where \texttt{[CLS]} and \texttt{[SEP]} are special tokens. \texttt{[EOU]} is the ``End Of Utterance" tag designed for multiturn context.  ${X}$ is then fed into the BERT encoder, which is a deep multi-layer bidirectional Transformer, to obtain a contextualized representation ${H}$. 


In detail, let ${X} = \{x_1, \dots, x_n\}$ be the embedding of the sequence, which are features of encoding sentence words of length $n$. The input embeddings are then fed into the multi-head attention layer to obtain the contextual representations.
	
The embedding sequence $X$ is processed to a multi-layer bidirectional Transformer for learning contextualized representations, which is defined as 
\begin{align}\label{eq:mutihead}
{H} = \textup{FFN}(\textup{MultiHead}(K,Q,V)),
\end{align}
where K,Q,V are packed from the input sequence representation $X$. As the common practice, we set $K=Q=V$ in the implementation.

For the following part, we use ${H} = \{h_1, \dots, h_n\}$ to denote the last-layer hidden states of the input sequence.

\subsection{SPIDER Training Objectives}
To simulate the dialogue-like features, we propose two pre-training objectives in addition to the original LM objectives: 1) utterance order restoration, which predicts the order of the permuted utterances in dialogue context; 2) sentence backbone regularization, which regularizes the model to improve the factual correctness of summarized subject-verb-object triplets. The utterance manipulations are shown in Figure \ref{struct}. The following subsections describe the objectives in turn.

\subsubsection{Utterance Order Restoration}\label{sec:uor}
Coherence is an essential aspect of conversation modeling. In a coherent discourse, utterances should respect specific orders of relations and logic. The ordering of utterances in a dialogue context determines the semantic of the conversation. Therefore, learning to order a set of disordered utterances in such a way that maximizes the discourse coherence will have a critical impact in learning the representation of dialogue contexts.

However, most previous studies focused on modeling the semantic relevance between the context and the response candidate. Here we introduce utterance-level position modeling, i.e., utterance order restoration to encourage the model to be aware of the semantic connections among utterances in the context. The idea is similar to autoencoding (AE) which aims to reconstruct the original data from corrupted input \cite{yang2019xlnet}. Given permuted dialogue contexts that comprise utterances in random orders, we maximize the expected log-likelihood of a sequence of the original ground-truth order. 

The goal of the utterance order restoration
is to organize randomly shuffled utterances of a conversation into a coherent dialogue context. We extract the hidden states of \texttt{[EOU]} from $H$ as the representation of each utterance. Formally, given an utterance sequence denoted as $C' = [H_{u_1}; H_{u_2}; \dots; H_{u_{K}}]$
with order $o = [o_1; o_2; \dots ; o_{K}]$, where $K$ means the number of maximum positions to be predicted. 
We expect an ordered context $C^{*} = [u_{o^{*}_1} ; u_{o^{*}_2} ; \dots; u_{o^{*}_{K}}]$ is the most coherent permutation of utterances.


As predicting the permuted orders is a more challenging optimization problem than NSP and SOP tasks due to the large searching space of permutations and causes slow convergence in preliminary experiments, we choose to only predict the order of the last few permuted utterances by a permutation ratio $\delta$ to control the maximum number of permutations: $K' = K*\delta$. The UOR training objective is then formed as:
\begin{equation}
\begin{split}
\mathbb{L}_{uor} &= -\sum_{k=1}^{K'}\left [ {o}_{k}\log\hat{o}_{k} \right ],
\end{split}
\end{equation}
where $\hat{o}_{k}$ denotes the predicted order.

\begin{figure*}
\centering
\includegraphics[width=1\textwidth]{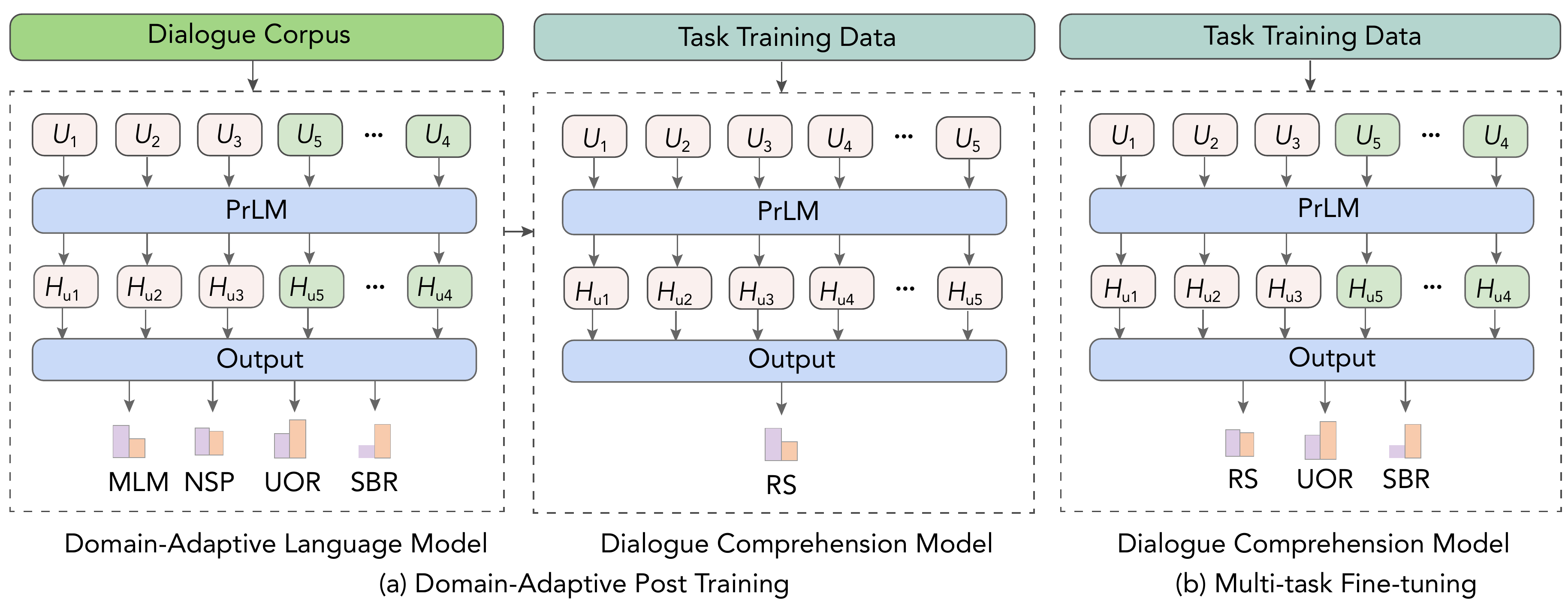}
\caption{Model training flow of domain-adaptive post-training (a) and multi-task fine-tuning (b).}
\label{model_training}
\end{figure*}

\subsubsection{Sentence Backbone Regularization}
The sentence backbone regularization objective is motivated to guide the model to distinguish the internal relation of the fact triplets that are extracted from each utterance, which would be helpful to improve the ability to capture the key facts of the utterance as well as the correctness. First, we apply a fact extractor to conduct the dependency parsing of each sentence. After that, we extract the subject, the root verb, and the object tokens as an SVO
triplet corresponding to each utterance. Inspired by \citet{bordes2013translating} where the embedding of the tail entity should be close to the embedding of the head entity plus some vector that depends on the relationship, we assume that given the dialogue input, in the hidden representation space, the summation of the subject
and the verb should be close to the object as
much as possible, i.e., 
\begin{equation}
    h_{subject} + h_{verb} \rightarrow h_{object}.\label{eq.svo}
\end{equation}

Consequently, based on the sequence hidden states
$h_i$ where $i = 1, ...,L_y$, we introduce a regularization
for the extracted facts:
\begin{equation}
    \mathbb{L}_{sbr} = \sum^{m}_{k=1}(1-\cos(h_{subj_k} + h_{verb_k}, h_{obj_k})),
\end{equation}
where $m$ is the total number of fact tuples extracted from the summary and $k$ indicates the $k$-th triplet.  ``$subj_k$", ``$verb_k$", and ``$obj_k$" are indexes of the $k$-th fact tuple's subject, verb, and object.

In our implementation, since PrLMs take subwords as input while the SVO extraction performs in word-level, we use the first-token hidden state as the representation of the original word following the way in \citet{devlin-etal-2019-bert} for named entity recognition.

\begin{table*}
	\centering\small
	 \resizebox{\linewidth}{!}
    {\setlength{\tabcolsep}{7.5pt}
		\begin{tabular}{cccccccccc}
			\toprule

			& \multicolumn{3}{c}{\textbf{Ubuntu}} & \multicolumn{3}{c}{\textbf{Douban}}  & \multicolumn{3}{c}{\textbf{ECD}}  \\
			&\textbf{Train} & \textbf{Valid} & \textbf{Test} & \textbf{Train} & \textbf{Valid} & \textbf{Test} & \textbf{Train} & \textbf{Valid} & \textbf{Test}  \\
						\midrule
			\midrule
			\# context-response pairs & 1M & 500K & 500K &1M & 50K & 10K  &1M & 10K & 10K\\
			\# candidates per context  & 2 & 10  & 10 & 2 & 2 & 10& 2 & 2 & 10\\
			Avg \#  turns per context & 10.13 & 10.11 & 10.11 & 6.69 & 6.75 & 6.45 & 5.51 & 5.48 & 5.64\\
			Avg \# words per utterance & 11.35 & 11.34 & 11.37 & 18.56 & 18.50 & 20.74 & 7.02 & 6.99 & 7.11\\
			\bottomrule
		\end{tabular}
	}
	\caption{\label{tab:dataset}  Data statistics of Ubuntu, Douban, and ECD datasets.}
\end{table*}

\section{Use of SPIDER Objectives}\label{sec:training}
In this section, we introduce two training methods to take the newly proposed language modeling objectives into account, namely domain-adaptive post-training and multi-task fine-tuning, as illustrated in Figure \ref{model_training}.
\subsection{Domain Adaptive Post-training}
Similar to BERT, we also adopt the masked language model (MLM) and the next sentence prediction (NSP) as LM-training tasks to enable our model to capture lexical and syntactic information from tokens in text.  More details of the LM training tasks can be found from \citet{devlin-etal-2019-bert}. The overall post-training loss is the sum of the MLM, NSP, UOR, and SBR loss.

Our full model is trained by a joint loss by combining both of the objectives above: \begin{equation}
    \mathbb{L} = \lambda_1(\mathbb{L}_{mlm}+\mathbb{L}_{nsp}) + \lambda_2\mathbb{L}_{uor}+\lambda_3\mathbb{L}_{sbr},
\end{equation}
where $\lambda_1, \lambda_2, \lambda_3$ are hyper-parameters.

After post-training the language model on the dialogue corpus, we load the pre-trained weights as the same way of using BERT \cite{devlin-etal-2019-bert}, to fine-tune the downstream tasks such as response selection and dialogue reasoning as focused in this work (details in Section \ref{sec:tasks}).

\subsection{Multi-task Fine-tuning}
Since our objectives can well share the same input as the downstream tasks, there is an efficient way of using multi-task fine-tuning (MTF) to directly train the task-specific models along with our SPIDER objectives. Therefore, we feed the permuted context to the dialogue comprehension model and combine the three losses for training:  
\begin{equation}
    \mathbb{L} = \beta_1\mathbb{L}_{dm} + \beta_2\mathbb{L}_{uor}+\beta_3\mathbb{L}_{sbr},
\end{equation}
where $\beta_1, \beta_2, \beta_3$ are hyper-parameters.

In order to train a task-specific model for dialogue comprehension, the hidden states $H$ will be fed into a classifier with a fully connected and softmax layer. We learn model $g(\cdot, \cdot)$ by minimizing cross entropy loss with dataset $\mathcal{D}$. Let $\Theta$ denote the parameters, for binary classification like the response selection task, the objective function $\mathcal{L(D}, \Theta)$ can be formulated as:
\begin{equation*}
\begin{split}
    \mathbb{L}_{dm} = -\sum_{i=1}^N [y_i\log(g(c_i,r_i)) + \\ (1-y_i)\log(1-g(c_i,r_i))].
\end{split}
\end{equation*}
where $N$ denotes the number of examples. For multiple choice task like MuTual, the loss function is:
\begin{equation*}
   \mathbb{L}_{dm} = -\sum_{i=1}^N\sum_{k=1}^C y_{i,c}\log(g(c_i,r_{i,k})).
\end{equation*}
where $C$ is the number of choice.

\section{Experiments}
\subsection{Datasets}\label{sec:tasks}
We evaluated our model on two English datasets: Ubuntu Dialogue Corpus (Ubuntu) \cite{lowe2015ubuntu} and Multi-Turn Dialogue Reasoning (MuTual) \cite{cui2020mutual},\footnote{Actually, MuTual is a retrieval-based dialogue corpus in form, but the theme is English listening comprehension exams, thus we regard as a reading comprehension corpus in this work. Because the test set of MuTual is not publicly available, we conducted the comparison with our baselines on the Dev set for convenience.} and two Chinese datasets: Douban Conversation Corpus (Douban) \cite{wu2016sequential} and E-commerce Dialogue Corpus (ECD) \cite{zhang2018modeling}. 
\subsubsection{Ubuntu Dialogue Corpus} Ubuntu \cite{lowe2015ubuntu} consists of English multi-turn conversations about technical support collected from chat logs of the Ubuntu forum. The dataset contains 1 million context-response pairs, 0.5 million for validation and 0.5 million for testing. In training set, each context has one positive response generated by human and one negative response sampled randomly. In validation and test sets, for each context, there are 9 negative responses and 1 positive response. 
\subsubsection{Douban Conversation Corpus} Douban \cite{wu2016sequential} is different from Ubuntu in the following ways. First, it is an open domain where dialogues are extracted from Douban Group. Second, response candidates on the test set are collected by using the last turn as the query to retrieve 10 response candidates and labeled by humans. Third, there could be more than one correct response for a context.
\subsubsection{E-commerce Dialogue Corpus} ECD \cite{zhang2018modeling} dataset is extracted from conversations between customer and service staff on Taobao. It contains over 5 types of conversations based on over 20 commodities. There are also 1 million context-response pairs in the training set, 0.5 million in the validation set, and 0.5 million in the test set.

\begin{table*}[t]
{
    \centering\small

    \resizebox{\linewidth}{!}
    {\setlength{\tabcolsep}{3pt}
        \renewcommand\arraystretch{1.1}
        \begin{tabular}{lcccccccccccc}
            \toprule \textbf{Model} & \multicolumn{3}{c}{\textbf{Ubuntu Corpus}} & \multicolumn{6}{c}{\textbf{Douban Conversation Corpus}} & \multicolumn{3}{c}{\textbf{E-commerce Corpus}} \\
            \cmidrule(r){2-4} \cmidrule(r){5-10} \cmidrule(r){11-13}
             &  $\textbf{R}_{10}$@1 & $\textbf{R}_{10}$@2 & $\textbf{R}_{10}$@5 & \textbf{MAP} & \textbf{MRR} & \textbf{P}@1 & $\textbf{R}_{10}$@1 & $\textbf{R}_{10}$@2 & $\textbf{R}_{10}$@5 & $\textbf{R}_{10}$@1 & $\textbf{R}_{10}$@2 & $\textbf{R}_{10}$@5 \\
            \midrule
            \midrule SMN  & 72.6 & 84.7 & 96.1 & 52.9 & 56.9 & 39.7 & 23.3 & 39.6 & 72.4 & 45.3 & 65.4 & 88.6 \\
             DUA  & 75.2 & 86.8 & 96.2 & 55.1 & 59.9 & 42.1 & 24.3 & 42.1 & 78.0 & 50.1 & 70.0 & 92.1 \\
             DAM  & 76.7 & 87.4 & 96.9 & 55.0 & 60.1 & 42.7 & 25.4 & 41.0 & 75.7 & - & - & - \\
             IoI  & 79.6 & 89.4 & 97.4 & 57.3 & 62.1 & 44.4 & 26.9 & 45.1 & 78.6 & - & - & - \\
             MSN & 80.0 & 89.9 & 97.8 & 58.7 & 63.2 & 47.0 & 29.5 & 45.2 & 78.8 & 60.6 & 77.0 & 93.7 \\
             MRFN & 78.6 & 88.6 & 97.6 & 57.1 & 61.7 & 44.8 & 27.6 & 43.5 & 78.3 & - & - & - \\
             SA-BERT & 85.5 & 92.8 & 98.3 & \textbf{61.9} & \textbf{65.9} & \textbf{49.6} & \textbf{31.3} & 48.1 & \textbf{84.7} & 70.4 & \textbf{87.9} & 98.5 \\
            \midrule
             \multicolumn{13}{l}{\textit{Multi-task Fine-tuning}} \\
              BERT &81.7 &90.4 &97.7 &58.8 &63.1 & 45.3 &27.7 &46.4 &81.8 & 61.7 & 81.1 & 97.0 \\
            \quad + SPIDER & 83.1 & 91.3  & 98.0 & 59.8 & 63.8  & 45.9 &28.5  & 48.7 & 82.6 & 62.6 & 82.7 & 97.1 \\
             \midrule
             \multicolumn{13}{l}{\textit{Domain Adaptive Post-training}} \\
             BERT & 85.7  &  93.0 & 98.5 &  60.5 & 64.7  & 47.4  & 29.1  & 47.8  & 84.9  & 66.4 & 84.8 & 97.6 \\
            \quad + SPIDER & \textbf{86.9} &  \textbf{93.8} &  \textbf{98.7} & 60.9 & 65.0  & 47.5 & 29.6 &  \textbf{48.8} & 83.6 & \textbf{70.8} & 85.3 &  \textbf{98.6} \\

             \bottomrule
        \end{tabular}
    }
    \caption{\label{tab:exp} Performance comparison on Ubuntu, Douban and E-Commerce datasets.
	}
}
\end{table*}

\subsubsection{Multi-Turn Dialogue Reasoning} MuTual \cite{cui2020mutual} consists of 8860 manually annotated dialogues based on Chinese student English listening comprehension exams. For each context, there is one positive response and three negative responses. The difference compared to the above three datasets is that only MuTual is reasoning-based. There are more than 6 types of reasoning abilities reflected in MuTual.
\subsection{Implementation Details}
For the sake of computational efficiency, the maximum number of utterances is specialized as 20. The concatenated context, response, \texttt{[CLS]} and \texttt{[SEP]} in one sample is truncated according to the ``longest first" rule or padded to a certain length, which is 256 for MuTual and 384 for the other three datasets. For the hyper-parameters, we
empirically set $\lambda_1 = \lambda_2 =\lambda_3 = \beta_1 = \beta_2 =1$ in our experiments.

Our model is implemented using Pytorch and based on the Transformer Library. We use BERT \cite{devlin-etal-2019-bert} as our backbone model. AdamW \cite{loshchilov2017decoupled} is used as our optimizer. The batch size is 24 for MuTual, and 64 for others. The initial learning rate is $4\times 10^{-6}$ for MuTual and $3\times 10^{-5}$ for others. The ratio is set to 0.4 in our implementation by default. We run 3 epochs for MuTual and 2 epochs for others and select the model that achieves the best result in validation. The training epochs are 3 for DAP. 

Our domain adaptive post-training for the corresponding response selection tasks is based on the three large-scale dialogue corpus including Ubuntu, Douban, and ECD, respectively.\footnote{Since phrases are quite common in Chinese, making it inaccurate to calculate the SVO relations according to Eq. \ref{eq.svo}, thus we did not use the SBR objective for the two Chinese tasks in this work.} The data statistics are in Table \ref{tab:dataset}. Since domain adaptive post-training is time-consuming, following  previous studies \cite{gu2020speaker}, we use  \textit{bert-base-uncased}, and \textit{bert-base-chinese} for the English and Chinese datasets, respectively. Because there is no appropriate domain data for the small-scale Mutual dataset, we only report the multi-task fine-tuning results with our SPIDER objectives, and also present the results with other PrLMs such as ELECTRA \cite{clark2020electra} for general comparison.

\subsection{Baseline Models}
We include the following models for comparison:

$\bullet$ \textbf{Multi-turn matching models}: Sequential Matching Network (SMN) \cite{wu2016sequential}, Deep Attention Matching Network (DAM) \cite{zhou2018multi}, Deep Utterance Aggregation (DUA) \cite{zhang2018modeling}, Interaction-over-Interaction (IoI) \cite{tao2019ioi} have been stated in Section \ref{sec:relatedwork}. Besides, Multi-Representation Fusion Network (MRFN) \cite{tao2019multi} matches context and response with multiple types of representations. Multi-hop Selector Network (MSN) \cite{yuan2019multi} utilizes a multi-hop selector to filter necessary utterances and matches among them.

$\bullet$ \textbf{PrLMs-based models}: BERT \cite{devlin2018bert}, SA-BERT \cite{gu2020speaker}, and ELECTRA \cite{clark2020electra}.

\subsection{Evaluation Metrics}
Following \cite{lowe2015ubuntu, wu2016sequential}, we calculate the proportion of true positive response among the top-$k$ selected responses from the list of $n$ available candidates for one context, denoted as $\textbf{R}_n$@$k$. Besides, additional conventional metrics of information retrieval are employed on Douban: Mean Average Precision (MAP) \cite{baeza1999modern}, Mean Reciprocal Rank (MRR) \cite{voorhees1999trec}, and precision at position 1 (P@1). 

\subsection{Results}
Tables \ref{tab:exp}-\ref{tab:mutual_result} show the results on the four benchmark datasets. We have the following observations:

1) Generally, the previous models based on multi-turn matching networks perform worse than simple PrLMs-based ones, illustrating the power of contextualized representations in context-sensitive dialogue modeling. PrLM can perform even better when equipped with our SPIDER objectives, verifying the effectiveness of dialogue-aware language modeling, where inter-utterance position information and inner-utterance key facts are better exploited. Compared with SA-BERT that involves more complex architecture and more parameters by injecting extra speaker-aware embeddings, SPIDER keeps the same model size as the backbone BERT, and even surpasses SA-BERT on most of the metrics.

2) In terms of the training methods, DAP generally works better than MTF, with the merits of two-step procedures including the pure LM-based post-training. According to the ablation study in Table \ref{tab:ablation}, we see that both of the dialogue-aware LM objectives are essentially effective and combining them (SPIDER) gives the best performance, which verifies the necessity of modeling the utterance order and factual correctness. We also notice that UOR shows better performance than SBR in DAP, while gives relative descent in MFT. The most plausible reason would be that UOR would permute the utterances in the dialogue context which helps the language model learn the utterance in UOR. However, in MFT, the major objective is the downstream dialogue comprehension task. The permutation of the context would possibly bring some negative effects to the downstream task training.

\begin{table}
{
    \centering\small
    \resizebox{\linewidth}{!}
    { \setlength{\tabcolsep}{8pt}
        \begin{tabular}{l l l l}
            \toprule 
            \textbf{Model}& MRR & $\textbf{R}_{4}$@1 & $\textbf{R}_{4}$@2  \\
            \midrule \midrule
             BERT$_{base}$  & 80.0 & 65.3 & 86.0 \\
              \quad + UOR & 80.7 & 66.1 & 86.7 \\
              \quad + SBR & 81.3 & 67.4 & 87.1 \\
              \quad + SPIDER & 81.6 & 67.6 & 87.3 \\
               \midrule
             BERT$_{large}$& 82.2& 69.1 & 87.9 \\
              \quad + UOR & 82.8& 69.8 & 88.6 \\
              \quad + SBR &83.4 & 71.0 &  89.4\\
              \quad + SPIDER &83.9 & 71.8 & 89.2 \\
               \midrule
             ELECTRA$_{base}$  &86.5 & 76.2 & 91.6 \\
              \quad + UOR & 86.9 & 76.6 & 91.8 \\
              \quad + SBR & 87.6 & 77.1 &  92.0 \\
              \quad + SPIDER &88.2 & 79.2 & 92.3 \\
               \midrule
             ELECTRA$_{large}$  & 94.9 & 90.6 & 97.7 \\
              \quad + UOR & 95.3 & 91.3 & 97.8 \\
              \quad + SBR & 95.5 & 91.6 & 97.8 \\
              \quad + SPIDER &  \textbf{95.6} &  \textbf{92.0} &  \textbf{97.9} \\
             \bottomrule
        \end{tabular}
    }
    \caption{\label{tab:mutual_result} Results on MuTual dataset.
	}
}
\end{table}



\begin{table}
		\centering\small
      \resizebox{\linewidth}{!}
    {  \setlength{\tabcolsep}{8pt}
		{
			\begin{tabular}{l l l l}
				\toprule
				\textbf{Model} & $\textbf{R}_{10}$@1 & $\textbf{R}_{10}$@2 & $\textbf{R}_{10}$@5 \\
				\midrule
				\midrule
				SPIDER$_\textup{DAP}$ &  \textbf{86.9} &   \textbf{93.8} &  \textbf{98.7} \\
				\quad w/o UOR & 86.2 & 93.3 & 98.6 \\ 
				\quad w/o SBR & 86.4 & 93.5  & 98.6 \\ 
				\quad w/o Both & 85.7  &  93.0 & 98.5 \\
				\midrule
				SPIDER$_\textup{MTF}$ &  83.1 & 91.3  & 98.0  \\ 
				\quad w/o UOR & 82.6 &  91.0 & 97.9   \\ 
				\quad w/o SBR & 82.3  & 90.8  & 97.8  \\ 
				\quad w/o Both & 81.7 &90.4 &97.7 \\
				\bottomrule
			\end{tabular}
		}
    }
		\caption{\label{tab:ablation} Ablation study on the Ubuntu dataset.}
	\end{table}

\subsection{Influence of Permutation Ratio}
For the UOR objective, a hyper-parameter $\delta$ is set to control the maximum number of permutations (as described in Section \ref{sec:uor}), which would possibly influence the overall model performance. To investigate the effect, we set the permutation ratio from [0, 20\%, 40\%, 60\%, 80\%, 100\%]. The result is depicted in Figure \ref{numImg}, in which our model outperforms the baseline in general, showing that the permutation indeed strengthens the baseline.

\subsection{Comparison with Different Context Length}
\begin{figure}
	\centering
			\setlength{\abovecaptionskip}{0pt}
			\begin{center}
			\pgfplotsset{height=5.3cm,width=6cm,compat=1.14,every axis/.append style={thick},every axis legend/.append style={at={(0.95,0.95)}},legend columns=3 row=1} \begin{tikzpicture} \tikzset{every node}=[font=\small] \begin{axis} [width=8cm,enlargelimits=0.13, legend pos=north west, xticklabels={0-4, 4-8,8-12,12-16,16-20}, axis y line*=left, axis x line*=left, xtick={1,2,3,4,5}, x tick label style={rotate=0},
			ylabel={${\rm R_{10} @1} $},
			ymin=83.5,ymax=90,
			ylabel style={align=left},xlabel={number of utterances},font=\small]
			\addplot+ [smooth, mark=*,mark size=1.2pt,mark options={mark color=cyan}, color=red] coordinates
			{(1, 85.2938347403428)  (2, 86.27162652589905)  (3,86.9946175991989)  (4,87.61989860583016)  (5,87.94123362906633)};
			\addlegendentry{\small SPIDER}
			\addplot+[smooth, mark=diamond*, mark size=1.2pt, mark options={mark color=cyan},  color=cyan] coordinates {(1, 84.3903300076746)  (2, 85.59551303200263)  (3,86.45568907247465)  (4,86.86797634136038)  (5,86.53523489932886)};
			\addlegendentry{\small Baseline}
			\end{axis}
			\end{tikzpicture}
		\end{center}
    	\caption{${\rm R_{10} @1} $ of SPIDER and the baseline BERT on different numbers of utterances.}
		\label{fig:num_utterance}
\end{figure}
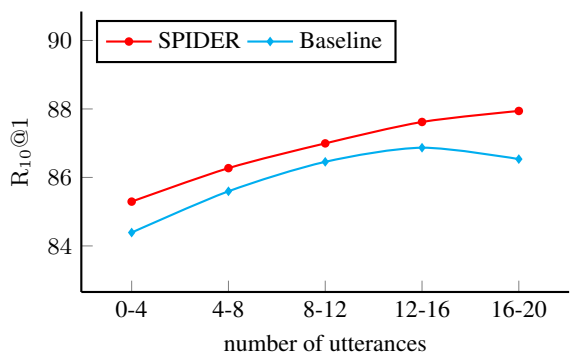

Context length can be measured by the number of turns and average utterance length in a conversation respectively. We split test instances from the Ubuntu dataset into several buckets and compare SPIDER with UOR with the BERT baseline. According to the results depicted in Figure \ref{fig:num_utterance}, we observe that SPIDER performs much better on contexts with long utterances, and it also performs robustly and is significantly and consistently superior to the baseline. The results indicate the benefits of modeling the utterance order for dialogue comprehension.

\subsection{Human Evaluation about Factual Correctness}
To compare the improvements of SPIDER over the baseline on factual correctness, we extract the error cases of the BERT baseline on MuTual (102 in total) and 42 (41.2\%) are correctly answered by SPIDER. Among the 42 solved cases, 33/42 (78.6\%) are entailed with SVO facts in contexts, indicating the benefits of factual correctness. 

\begin{figure}
\setlength{\abovecaptionskip}{0pt}
\pgfplotsset{height=5.3cm,width=8cm,compat=1.15,every axis/.append style={thick},every axis legend/.append style={at={(0.95,0.95)}},legend columns=3 row=1} \begin{tikzpicture} \tikzset{every node}=[font=\small]
\begin{axis} [width=8cm,enlargelimits=0.13,legend pos=north west,xticklabels={0.0,0.2,0.4,0.6,0.8,1.0}, axis y line*=left, axis x line*=left, xtick={0,1,2,3,4,5}, x tick label style={rotate=0},
  ylabel={$\textbf{R}_{10}$@1}, ymin=85.4,ymax=87.8,
  ylabel style={align=left},font=\small]
+\addplot+ [smooth, mark=*,mark size=1.2pt,mark options={mark color=red}, color=red] coordinates { (0,85.7) (1,86.21) (2,86.9) (3,86.7) (4,86.4) (5,86.38)};
\addlegendentry{\small Our method}
\addplot+[densely dotted, mark=none, color=cyan] coordinates {(0, 85.7)(1, 85.7)(2, 85.7)(3, 85.7)(4, 85.7)(5, 85.7)};
\addlegendentry{\small Baseline}
\end{axis}
\end{tikzpicture}
     \caption{\label{numImg}Influence of the permutation ratio $\delta$.}
\end{figure}
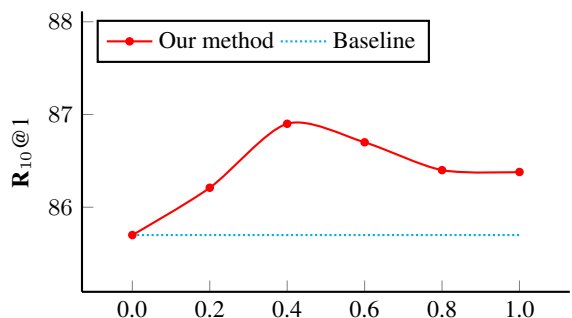

\section{Conclusion}
In this paper, we focus on the task-related adaptation of the pre-trained language models and propose SPIDER (Structural Pre-traIned DialoguE Reader), a structural language modeling method to capture dialogue exclusive features. To explicitly model the coherence among utterances and the key facts in each utterance, we introduce two novel dialogue-aware language modeling tasks including utterance order restoration and sentence backbone regularization objectives. Experiments on widely-used multi-turn dialogue comprehension benchmark datasets show the superiority over baseline methods. Our work reveals a way to make better use of the structure learning of the contextualized representations from pre-trained language models and gives insights on how to adapt the language modeling training objectives in downstream tasks.

\bibliographystyle{acl_natbib}
\bibliography{anthology,acl_output}


\end{document}